%% file: main.tex
\documentclass[runningheads]{llncs}

\usepackage{eccvabbrv}

\usepackage{graphicx}
\usepackage{booktabs}
\usepackage{wrapfig}
\usepackage[accsupp]{axessibility}  % Improves PDF readability for those with disabilities.

\usepackage{hyperref}

\usepackage{orcidlink}
\usepackage{subcaption}
\usepackage{cleveref}
\begin{document}

\title{DiT as Real-Time Rerenderer: Streaming Video Stylization with Autoregressive Diffusion Transformer} 

\titlerunning{DiT as Real-Time Rerenderer}

\author{
Hengye Lyu\inst{1}\orcidlink{0009-0009-0549-5077} \and
Zisu Li\inst{2}\orcidlink{0000-0001-8825-0191} \and
Yue Hong\inst{1}\orcidlink{0009-0001-9940-1650} \and
Yueting Weng\inst{3}\orcidlink{0000-0003-1649-8221} \and
Jiaxin Shi\inst{3}\orcidlink{0000-0003-0065-8374} \and
Hanwang Zhang\inst{4}\orcidlink{0000-0001-7374-8739} \and
Chen Liang\inst{1}\orcidlink{0000-0003-0579-2716}
}

\authorrunning{H. Lyu et al.}

\institute{
The Hong Kong University of Science and Technology (Guangzhou)\\
\email{hengyelyu@gmail.com; yhong252@connect.hkust-gz.edu.cn; chenliang2@hkust-gz.edu.cn}
\and
The Hong Kong University of Science and Technology\\
\email{vespersue98@gmail.com}
\and
XMax.AI Ltd.
\email{\{yueting,jiaxin\}@xmax.ai}
\and
Nanyang Technological University
\email{hanwangzhang@ntu.edu.sg}
}

\maketitle

\begin{figure}[h]
  \centering
  \includegraphics[height=0.5\textwidth]{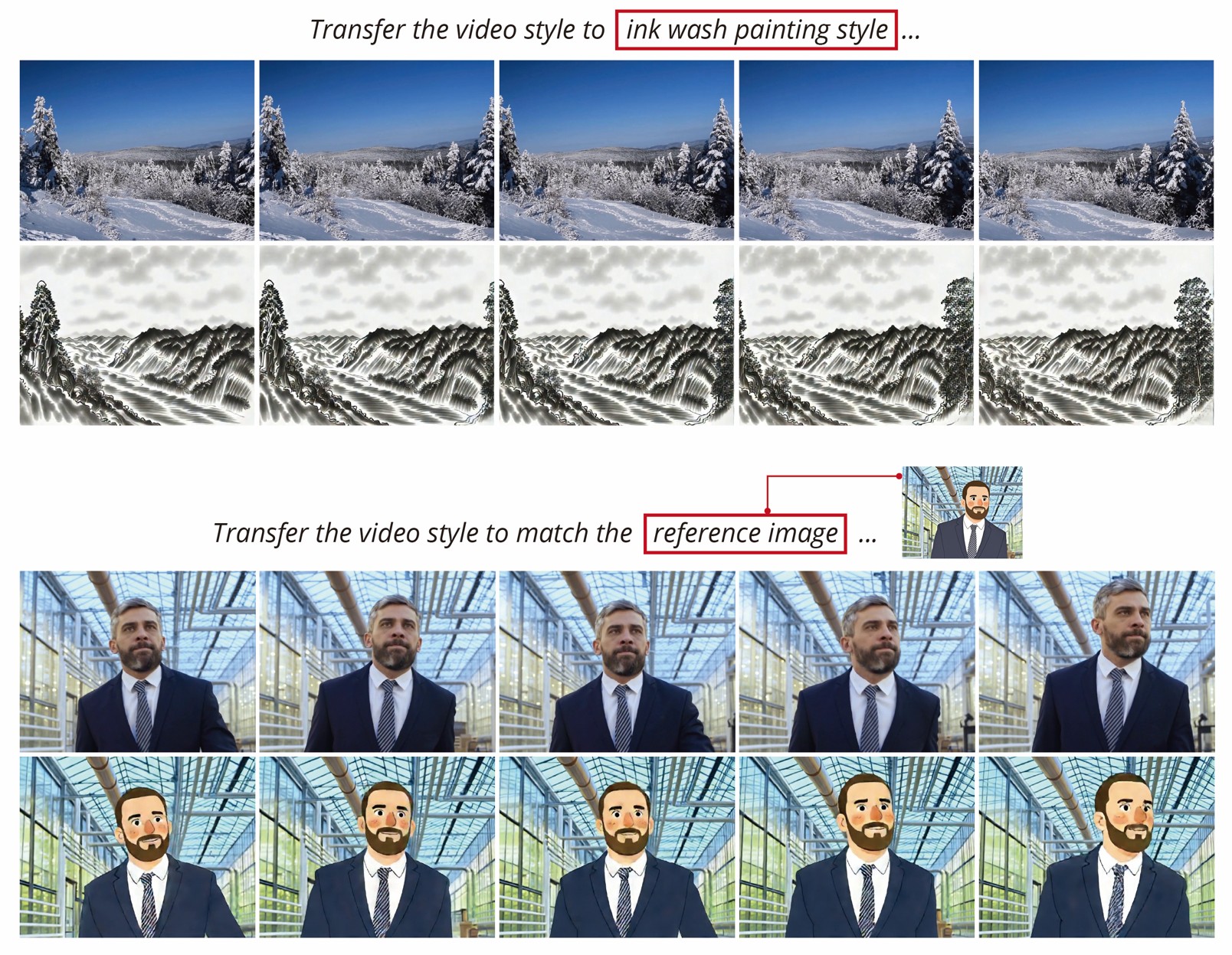}
  \caption{RTR-DiT is a DiT-based real-time video stylization framework that enables streaming rerendering of video frames, producing stable and consistent results. Top: Text-guided video-to-video (TV2V) stylization results.Bottom: Reference-guided video-to-video (RV2V) stylization results.}
  \label{fig:demo}
\end{figure}

\input{0_abstract}
\input{1_introduction}

\input{2_related_work}
\input{3_method}
\input{4_experiments}

\input{5_conclusion}

\bibliographystyle{splncs04}
\bibliography{main}
\end{document}

%% file: 0_abstract.tex
\begin{abstract}
Recent advances in video generation models has significantly accelerated video generation and related downstream tasks. Among these, video stylization holds important research value in areas such as immersive applications and artistic creation, attracting widespread attention. However, existing diffusion-based video stylization methods struggle to maintain stability and consistency when processing long videos, and their high computational cost and multi-step denoising make them difficult to apply in practical scenarios.
In this work, we propose RTR-DiT (DiT as Real-Time Rerenderer), a steaming video stylization framework built upon Diffusion Transformer. We first fine-tune a bidirectional teacher model on a curated video stylization dataset, supporting both text-guided and reference-guided video stylization tasks, and subsequently distill it into a few-step autoregressive model via post-training with Self Forcing and Distribution Matching Distillation. Furthermore, we propose a reference-preserving KV cache update strategy that not only enables stable and consistent processing of long videos, but also supports real-time switching between text prompts and reference images. 
Experimental results show that RTR-DiT outperforms existing methods in both text-guided and reference-guided video stylization tasks, in terms of quantitative metrics and visual quality, and demonstrates excellent performance in real-time long video stylization and interactive style-switching applications.

  \keywords{Video Stylization \and Autoregressive DiT \and Real-Time Video Generationl}
\end{abstract}

%% file: 1_introduction.tex
\section{Introduction}
\label{sec:intro}

In recent years, video diffusion models (VDMs) \cite{blattmann2023svd,ho2022vdm,peebles2023dit} have achieved remarkable progress in high-quality video generation, profoundly advancing research in computer vision and generative artificial intelligence. Leveraging pretrained Text-to-Video (T2V) and Image-to-Video (I2V) diffusion models\cite{yang2024cogvideox,wan2025}, researchers have further explored a variety of downstream tasks. Among these tasks, video stylization, which aims to transfer a target visual style to an input video while preserving its underlying content and temporal coherence, has attracted increasing attention due to its wide applications in digital content creation, film production, animation, and personalized media generation.

Video stylization based on diffusion models has been extensively researched over a long period. Early methods based on Latent Diffusion Models (LDMs) \cite{rombach2022ldm} extended image stylization to video scenes. These methods can generally be divided into two technical approaches: one maps the video into latent space using techniques like DDIM inversion\cite{song2020ddim} and injects style features, while the other extracts structural and motion information from the scene using frameworks like ControlNet\cite{zhang2023controlnet}, and rerenders based on these priors. Additionally, these methods typically stylize key frames and employ frame propagation strategies to handle long videos \cite{geyer2023tokenflow,jeong2023ground-a-video,yang2023rerender-a-video,yang2024fresco,zhu2025tokenwarping}. However, LDM-based methods face issues with frame-to-frame inconsistency, and their approach to handling long videos limits their ability to be applied in real-time scenarios. Recently, Diffusion Transformer (DiT)-based methods have significantly outperformed LDM-based methods in terms of generation quality and consistency. Some works have extended DiT to video stylization tasks by fine-tuning the DiT blocks\cite{ye2025stylemaster}. Moreover, some works have designed All-In-One video editing frameworks\cite{jiang2025vace,wu2025insvie} with stylization as a subtask. However, the 3D full attention in DiT blocks is bidirectional in temporal modeling, meaning each frame's estimation depends on both past and future frames, making it challenging to scale to real-time and long video applications.

Recently, many studies have focused on accelerating diffusion models, making DiT-based models viable for real-time applications. Among these, Distribution Matching Distillation (DMD)\cite{yin2024dmd,yin2024dmd2} distills multi-step sampling diffusion models into single-step or few-step generators, significantly improving generation efficiency while maintaining high output quality. Additionally, some methods have achieved autoregressive DiT models by incorporating causal masks\cite{yin2025causvid}, and used techniques such as Diffusion Forcing\cite{chen2024diffusion-forcing} and Self Forcing\cite{huang2025self-forcing} to reduce cumulative errors, thereby enhancing the stability and quality of long-video inference. These methods contribute to real-time video generation applications and further extend to Video-to-Video tasks. 

In this paper, we propose RTR-DiT (DiT as Real-Time Rerenderer), a real-time video stylization framework built upon an autoregressive DiT. By leveraging the strong generative priors of pretrained video diffusion models, RTR-DiT inherits their powerful capability for high-quality video synthesis, enabling temporally coherent and visually consistent stylized video generation. Moreover, our approach advances video stylization beyond traditional methods that primarily focus on transferring local texture or geometric patterns. Instead, RTR-DiT enables semantically consistent and high-fidelity style transformation, allowing complex visual styles to be applied at the scene and object level while preserving the underlying semantic structure and motion dynamics of the original video. As shown in \cref{fig:demo}, RTR-DiT supports both text and reference images as style guidance conditions, processing video frames in a streaming manner to generate high-quality and consistent stylized results.

To implement RTR-DiT, we first constructed a long video stylization dataset containing text-video and reference-video pairs to support the training of our model. Next, we designed a video stylization training and inference framework, which uses text and reference images as conditions to transform the input video into a stylized version. During training, we first fine-tuned a multi-step DiT with temporal bidirectional modeling as the teacher model. Then, in the post-training stage, we distilled it into an autoregressive few-step generator by combining Self Forcing and DMD techniques, enabling real-time stylization for videos. To ensure stable processing of long videos, we designed a reference-preserving KV cache update strategy to improve the consistency of stylization, while also supporting real-time switching between text and reference image conditions, making it more suitable for interactive applications.

In summary, the main contributions of this work are as follows:

\begin{itemize}

  \item We propose RTR-DiT, the first real-time video stylization framework built on DiT. Conditioned on text prompts and reference images, RTR-DiT performs streaming video-to-video stylization while preserving high visual fidelity and temporal coherence.

  % \item We develop a training and inference pipeline for a few-step causal generator based on DiT. Specifically, we distill a multi-step teacher with bidirectional temporal modeling into a few-step streaming stylization model, enabling efficient real-time inference and stable stylization over long video sequences.

  \item We propose a reference-preserving KV cache update strategy that ensures the autoregressive DiT can stably process video frames during long video stylization, resulting in continuous and condition-consistent stylized outputs.

  \item We propose an interactive video style-switching method that modifies the video style in real-time by changing the guidance conditions, and extend it to applications in real-world scenarios.

\end{itemize}\textbf{}

%% file: 2_related_work.tex
\section{Related Work}

\subsubsection{Video Stylization via Diffusion Models.}
The success of diffusion models in image stylization has motivated researchers to extend them to the video domain. Some representative works employ U-Net networks to transform video frames in the latent space using text or reference frames\cite{guo2023animatediff,ku2024anyv2v,singer2022make-a-video,wu2023tune-a-video,kara2024rave}. However, these methods often face challenges in maintaining temporal consistency between frames, and thus typically require additional constraints, such as optical flow \cite{geyer2023tokenflow,cong2023flatten,jeong2023ground-a-video}. Meanwhile, network architectures such as ControlNet\cite{zhang2023controlnet} enable generation conditioned on scene structures, motion, and other features, motivating several works to explore ControlNet-based video stylization methods\cite{yang2024fresco,zhu2025tokenwarping}.

 Recently, many video stylization works have been built upon the DiT model. Several works directly train DiT networks to achieve end-to-end style transfer, such as StyleMaster\cite{ye2025stylemaster} and DreamStyle\cite{li2026dreamstyle}. Other works have explored video editing frameworks, including stylization\cite{wu2025insvie,jiang2025vace}. Although these methods achieve significant improvements in video quality, they still face challenges in handling long videos and in real-time scenarios.

Furthermore, existing commercial models\cite{gen-4-aleph,kling} only offer video editing APIs with limited duration constraints, failing to meet the demands of long video processing. In contrast, our method can handle longer-duration video editing and supports interactive applications.

\subsubsection{Real-Time Video Diffusion Models.} 

Real-time video diffusion models enable low-latency streaming by factorizing synthesis along the temporal dimension. Some methods, such as CogVideo\cite{yang2024cogvideox}, VideoPoet\cite{kondratyuk2023videopoet}, and MAGI\cite{teng2025magi}, adopt autoregressive modeling with spatio-temporal tokens. Although this approach supports streaming generation, purely autoregressive models often struggle to balance visual fidelity and long-range temporal coherence\cite{chen2024diffusion-forcing,yin2025causvid}. 
Moreover, autoregressive video generation introduces a fundamental challenge in long rollouts due to the training–test mismatch caused by Teacher Forcing, which leads to exposure bias and error accumulation over time. Several forcing paradigms like Diffusion Forcing\cite{chen2024diffusion-forcing} and Self Forcing\cite{huang2025self-forcing} have been proposed to address this issue. Diffusion Forcing perturbs temporal context by assigning different noise levels to frames and denoising them jointly, improving robustness to corrupted histories in streaming settings\cite{yin2025causvid}. Self Forcing further aligns training with inference by conditioning models on self-generated rollouts during training. In parallel, efficiency-oriented techniques such as few-step solvers and distillation-based generators\cite{zhang2023controlnet,luo2023latent-consistency-models,yin2024dmd,yin2024dmd2} reduce inference latency, but are primarily validated in offline or image-centric scenarios and remain challenging to integrate reliably with long-duration streaming video diffusion. Despite this progress, existing real-time video diffusion models are largely designed for generation rather than video-to-video translation, and struggle to maintain stability and reference consistency under long-horizon streaming inputs. These limitations motivate approaches that explicitly reformulate video stylization as a streaming process with causal temporal modeling and efficient inference. 

%% file: 3_method.tex
\section{Method}

\begin{figure}[t]
  \centering
  \includegraphics[width=0.8\textwidth]{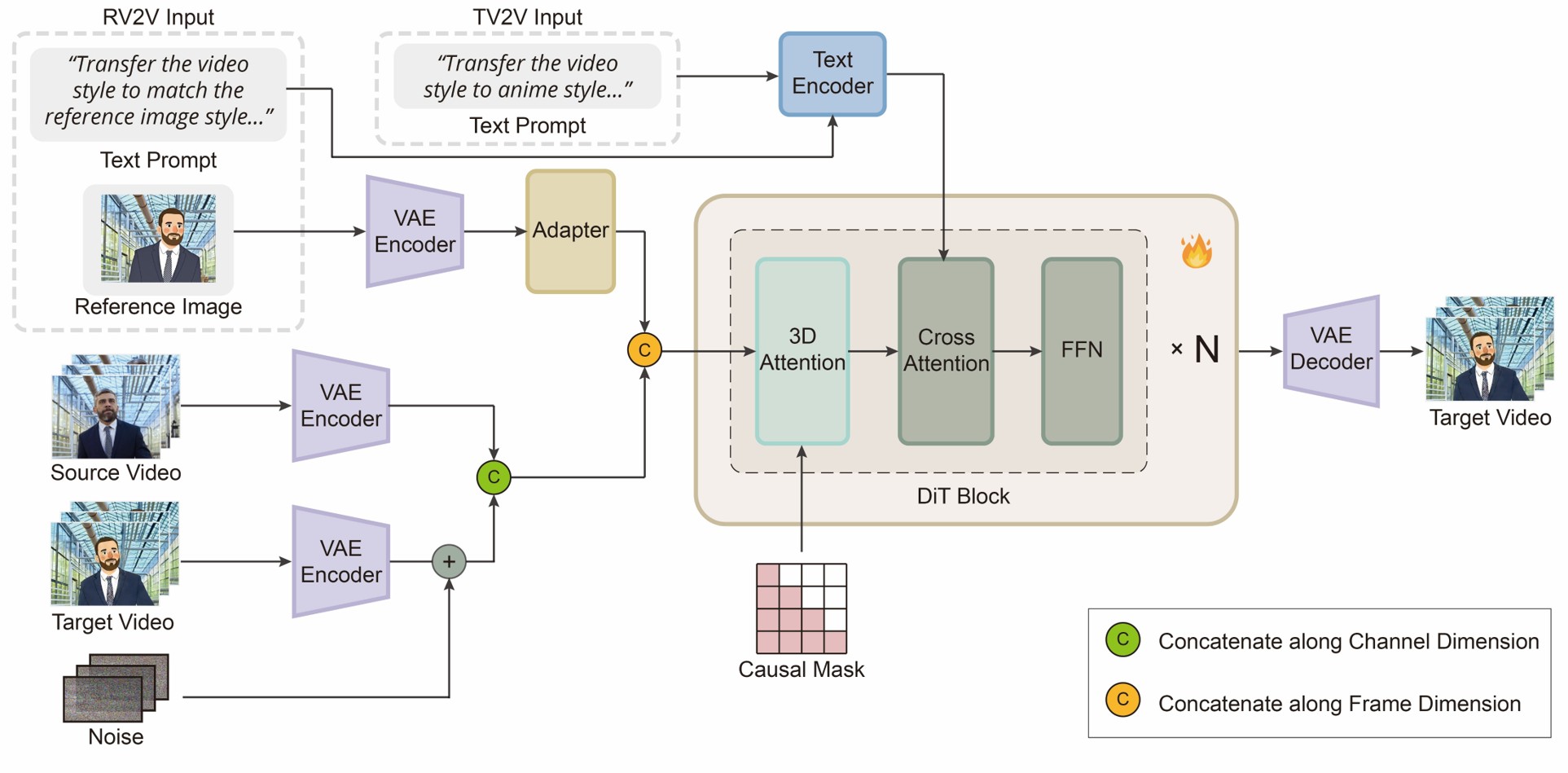}
  \caption{Overview of RTR-DiT. During training, the original video, the target video, and the reference image are first encoded using a VAE. The noised target video is then concatenated with the source video along the feature dimension, and further concatenated with the reference image along the temporal dimension. When training the Causal DiT, a causal mask is applied within the 3D attention mechanism, enabling the model to perform autoregressive prediction.}
  \label{fig:overview}
\end{figure}

RTR-DiT is an autoregressive DiT model that enables real-time streaming stylization of video frame sequences, guided by prompts or reference images. We refer to this process as "rerendering", which involves editing the appearance and style of the content while preserving the original scene structure and intrinsic attributes. This process includes two modes: Text-guided video-to-video (TV2V) stylization and Reference-guided video-to-video (RV2V) stylization.

\subsection{Dataset Creation for Video Stylization}
Currently, there is a lack of high-quality, long-duration video stylization datasets that support both text and reference guidance. To address this, we construct a training dataset using existing commercial and open-source large models. We begin by collecting 5,000 videos from the internet, covering a wide range of scenes including human-centric content, animals, and natural scenes. Next, we use Qwen3-VL\cite{bai2025qwen3-vl} to generate multiple stylized text prompts and then employ Kling\cite{kling} to obtain the corresponding stylized videos. For TV2V mode, we use the prompts Qwen3-VL generated as the text condition. For RV2V mode, we randomly select a frame from the target video as the reference and fix the guiding text to "\emph{Transfer the video style to match the reference image style …}"

\subsection{Preliminary of Base Model}

We use the DiT-based Wan\cite{wan2025} video model as the base model. As shown in \cref{fig:overview}, During training, we encode the source video $V_s$ and the target video $V_t$ with a variational autoencoder (VAE)\cite{wan2025}, obtaining latent representations for $N$ frames, $\mathbf{x}^{1:N}=(\mathbf{x}^1,\mathbf{x}^2,\ldots,\mathbf{x}^N)$ and $\mathbf{z}_0^{1:N}=(\mathbf{z}_0^1,\mathbf{z}_0^2,\ldots,\mathbf{z}_0^N)$, respectively. 
If a reference image $I_r$ is provided, we also obtain its VAE latent $\mathbf{c}_{\text{ref}}$. 
The text prompt is encoded by a text encoder \cite{raffel2020t5} to produce the text conditioning embedding $\mathbf{c}_{\text{text}}$.

Following prior work \cite{yin2025causvid,huang2025self-forcing}, we first fine-tune a bidirectional DiT as a teacher for our autoregressive DiT. 
Given a clean latent $\mathbf{z}_0$, we perturb it with noise at a continuous timestep $t\in[0,1]$ to obtain the noisy latent $\mathbf{z}_t$. 
We then concatenate $\mathbf{z}_t$ with the source video latent $\mathbf{x}$ along the channel dimension. 
The source latent $\mathbf{x}$ encodes content and structure cues from the original video, including spatial layout and motion patterns, and provides structural guidance that helps rerendering preserve scene geometry and temporal coherence.

The concatenated latent is then patchified into a sequence of tokens. In RV2V mode, the reference latent $\mathbf{c}_{\text{ref}}$ is processed by a lightweight adapter and prepended to the token sequence. All tokens are subsequently fed into the DiT blocks for modeling.

We fine tune the model using a flow matching\cite{lipman2022flow-matching} objective:
\begin{equation}
\mathcal{L}_{\text{FM}}
=
\mathbb{E}_{\mathbf{z}_0, \mathbf{z}_1, c, t}
\left\|
\mathbf{v}_\theta(\mathbf{z}_t, \mathbf{x}, c, t)
-
\left(\mathbf{z}_1-\mathbf{z}_0\right)
\right\|_2^2,
\label{eq:fm}
\end{equation}
where $\mathbf{z}_0$ denotes the clean latent encoded from data, $\mathbf{z}_1 \sim \mathcal{N}(0,\mathbf{I})$ is sampled from the standard Gaussian, and
\begin{equation}
\mathbf{z}_t = (1-t)\mathbf{z}_0 + t\mathbf{z}_1,\quad t\sim \mathcal{U}(0,1).
\end{equation}
Here $\mathbf{v}_\theta$ denotes the predicted velocity field, and the conditioning $c$ is defined as $c=\{\emptyset,\mathbf{c}_{\text{text}}\}$ for TV2V, or $c=\{\mathbf{c}_{\text{ref}},\mathbf{c}_{\text{text}}\}$ for RV2V.

\subsection{Autoregressive DiT for Streaming Stylization}
\begin{figure}[t]
  \centering
  \includegraphics[width=1\textwidth]{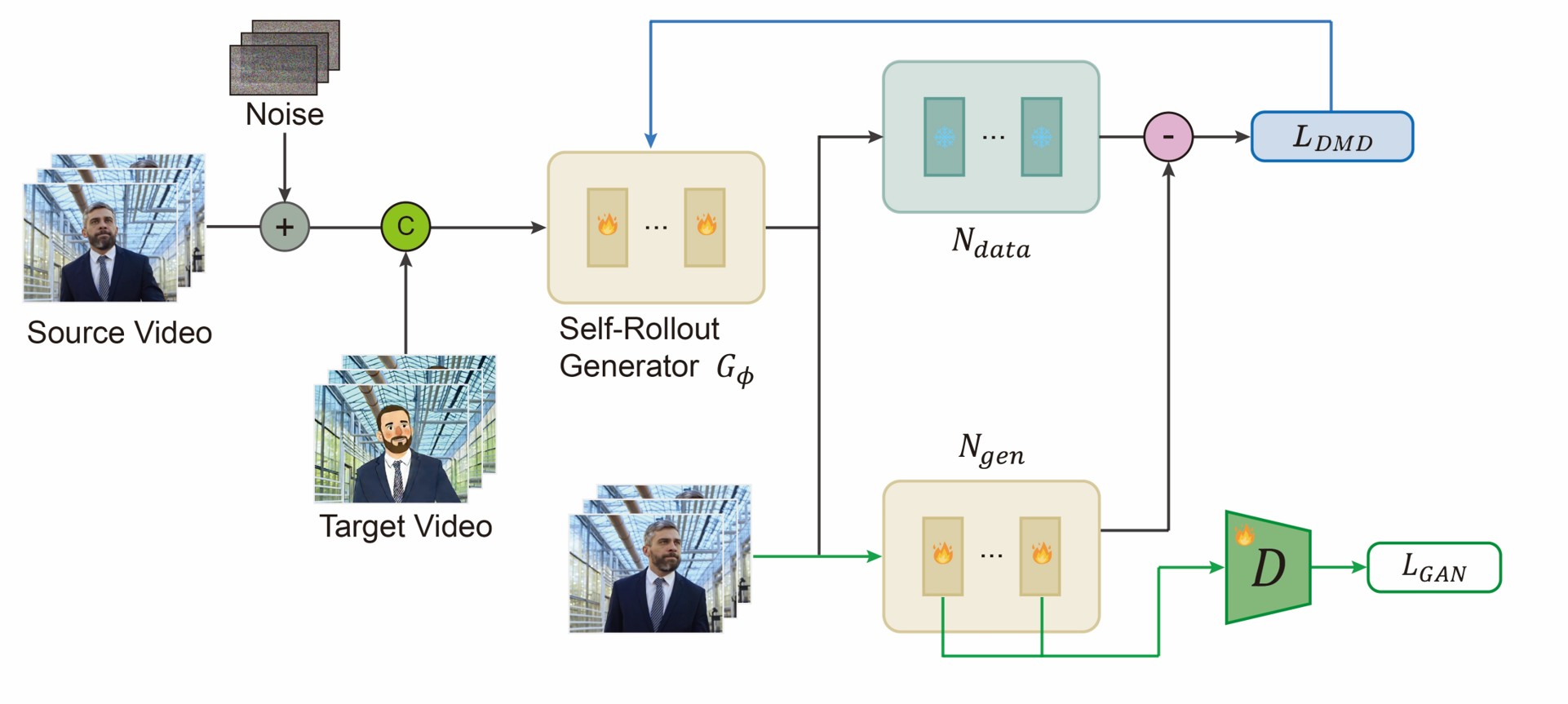}
  \caption{Post-training pipeline. We distill the bidirectional teacher model into a few-step autoregressive generator using Self-Forcing and DMD, and further incorporate adversarial training to improve generation performance.}
  \label{fig:pipeline}
\end{figure}

In the teacher model, each block uses full 3D spatiotemporal attention with bidirectional temporal modeling, meaning that the prediction of each frame depends on both past and future frames. This bidirectional dependency is incompatible with streaming video editing. To address this, we convert DiT into an autoregressive form to enable streaming inference and frame-by-frame generation.

Specifically, we impose a temporal causal mask on the 3D attention so that tokens in the noisy latent frame $\mathbf{z}_t^{n}$ can only attend to tokens from the current frame and preceding frames, \ie, $\{\mathbf{z}_t^{1},\mathbf{z}_t^{2},\ldots,\mathbf{z}_t^{n}\}$.
Let $Q$, $K$, and $V$ be the query, key, and value matrices computed from the token sequence of $\mathbf{z}_t$, the masked attention is
\begin{equation}
\begin{aligned}
\mathrm{Attention}(Q,K,V)
&= \mathrm{softmax}\!\left(
\frac{QK^\top}{\sqrt{d}} + M
\right)V, \\
\text{where}\quad
M^{ij} &=
\begin{cases}
0, & \left\lfloor j/k \right\rfloor \le \left\lfloor i/k \right\rfloor, \\
-\infty, & \text{otherwise}.
\end{cases}
\end{aligned}
\label{eq:causal-attn}
\end{equation}
Here $M$ is the causal mask, $d$ is the attention head dimension, and $k$ denotes the number of spatial tokens per latent frame. This construction ensures that each token attends only to tokens from the current and past frames.

Moreover, to improve generation efficiency and enable few-step inference, we perform denoising along a fixed timestep sequence $\{s_1, s_2, \cdots, s_K\}$, where $s_i\in t$, instead of iterating densely over the entire time horizon. Under this setting, we obtain a few-step autoregressive generator $G_\phi$, which is further optimized through post-training.

\subsubsection{Self-Rollout.}

We adopt Self Forcing training, where the model conditions each frame generation step on its own previously generated frames rather than ground-truth history, thereby reducing the train–test gap in autoregressive inference. Formally, the conditional generation distribution of $G_\phi$ under Self Forcing is
\begin{equation}
    p_{\phi}\!\left(\hat{\mathbf{z}}_0^{1:N}\mid \mathbf{x}^{\le N}, c\right)
    =
    \prod_{n=1}^{N}
    p_{\phi}\!\left(\hat{\mathbf{z}}_0^{n}\mid \hat{\mathbf{z}}_0^{<n}, \mathbf{x}^{\le n}, c\right).
\end{equation}

\subsubsection{Distribution Matching Distillation.}

We adopt Distribution Matching Distillation (DMD)\cite{yin2024dmd,yin2024dmd2} to distill the bidirectional, multi-step teacher model into $G_\phi$. As shown in \cref{fig:pipeline}, we initialize two score networks with the teacher’s weights: a frozen network $N_{\text{data}}$ for estimating the score of the real data distribution, and a trainable network $N_\text{gen}$ for estimating the score of the generator-induced distribution. DMD aims to minimize the reverse Kullback-Leibler (KL) divergence between the real and generated distributions. Its gradient can be expressed in terms of two score functions as follows:

\begin{equation}
\begin{aligned}
\nabla_\phi\mathcal{L}_{\text{DMD}} = \mathbb{E}_{\mathbf{z}_t,\mathbf{x},c,t} \Big[ - \big( 
  & S_{\text{data}}\big(F(G_\phi(\mathbf{z}_t,\mathbf{x},c),t),t\big) \\
  & - S_{\text{gen}}\big(F(G_\phi(\mathbf{z}_t,\mathbf{x},c),t),t\big) \big) \frac{dG}{d\phi} \Big]
\end{aligned}
\label{eq:dmd}
\end{equation}
where $F$ is forward diffusion process, and $S_{\text{data}}$ and $S_{\text{gen}}$ are score functions\cite{song2020score-SDE} estimated by $N_{\text{data}}$ and $N_{\text{gen}}$, respectively. After each update of $G_\phi$, we update $\mu_{\text{gen}}$ five times to match the score of the current generator distribution\cite{yin2025causvid}. 

\subsubsection{Adversarial Post-Training.} 
Moreover, we adopt adversarial training \cite{lin2025diff-adv,yin2024dmd2} to further improve the generation quality of $G_\phi$. We attach an additional classification branch to $N_{\text{gen}}$ as a discriminator $D$, and optimize the $G_\phi$ and $D$ using a standard GAN objective:
\begin{equation}
\label{eq:gan}
\begin{aligned}
\mathcal{L}_{\text{GAN}}
&=
\mathbb{E}_{\mathbf{z}_0,\mathbf{x},c}\!\left[\log D\!\left(\mathbf{z}_0,\mathbf{x}, c\right)\right]
+\mathbb{E}_{\mathbf{z}_t,\mathbf{x},c,t}\!\left[-\log D\!\left(F(G_{\phi}(\mathbf{z}_t,\mathbf{x},c),t),\mathbf{x},c\right)\right]
\end{aligned}
\end{equation}

\subsection{Inference for Consistent Long-Video Generation}

To enable continuous editing of arbitrarily long videos, we follow the approach of Huang \etal~\cite{huang2025self-forcing} and adopt a rolling KV cache mechanism. Specifically, we initialize fixed-length KV caches for both 3D attention and cross attention. During inference, newly generated frames are appended to the KV cache of the 3D attention, and the cache is updated in a rolling manner. Once the cache reaches its maximum capacity, the oldest frames are removed to make room for new ones.

In RV2V mode, a simple strategy is to prepend the reference image before the first frame and update the KV cache in a rolling manner. However, once the cache reaches its maximum capacity, the reference image is removed, preventing subsequent frames from attending to it and leading to noticeable degradation.

To address this issue, we propose a reference-preserving KV cache update strategy. After the cache is fully populated, the reference image is permanently anchored at the front of the cache, while the remaining positions continue to roll forward as new frames are generated. This design ensures that all subsequent frames can consistently attend to the reference image, thereby enabling stable long-video generation.

Furthermore, we modify the KV cache update mechanism to enable real-time switching of video styles by changing the text prompt or reference image. When a new text prompt or reference image is provided to the ongoing stylization model, the KV cache of the cross attention is reinitialized. Meanwhile, the 3D attention KV cache retains the most recent frame to ensure a smooth transition during the style change. If the new input is a reference image, it is anchored at the front of the KV cache so that subsequent frames can consistently attend to the reference style. This strategy enables more flexible deployment in real-time interactive scenarios.

%% file: 4_experiments.tex
\section{Experiments}
\subsection{Experimental Settings}
\subsubsection{Training Details.}

\begin{figure}[t]
  \centering
  \includegraphics[width=1\textwidth]{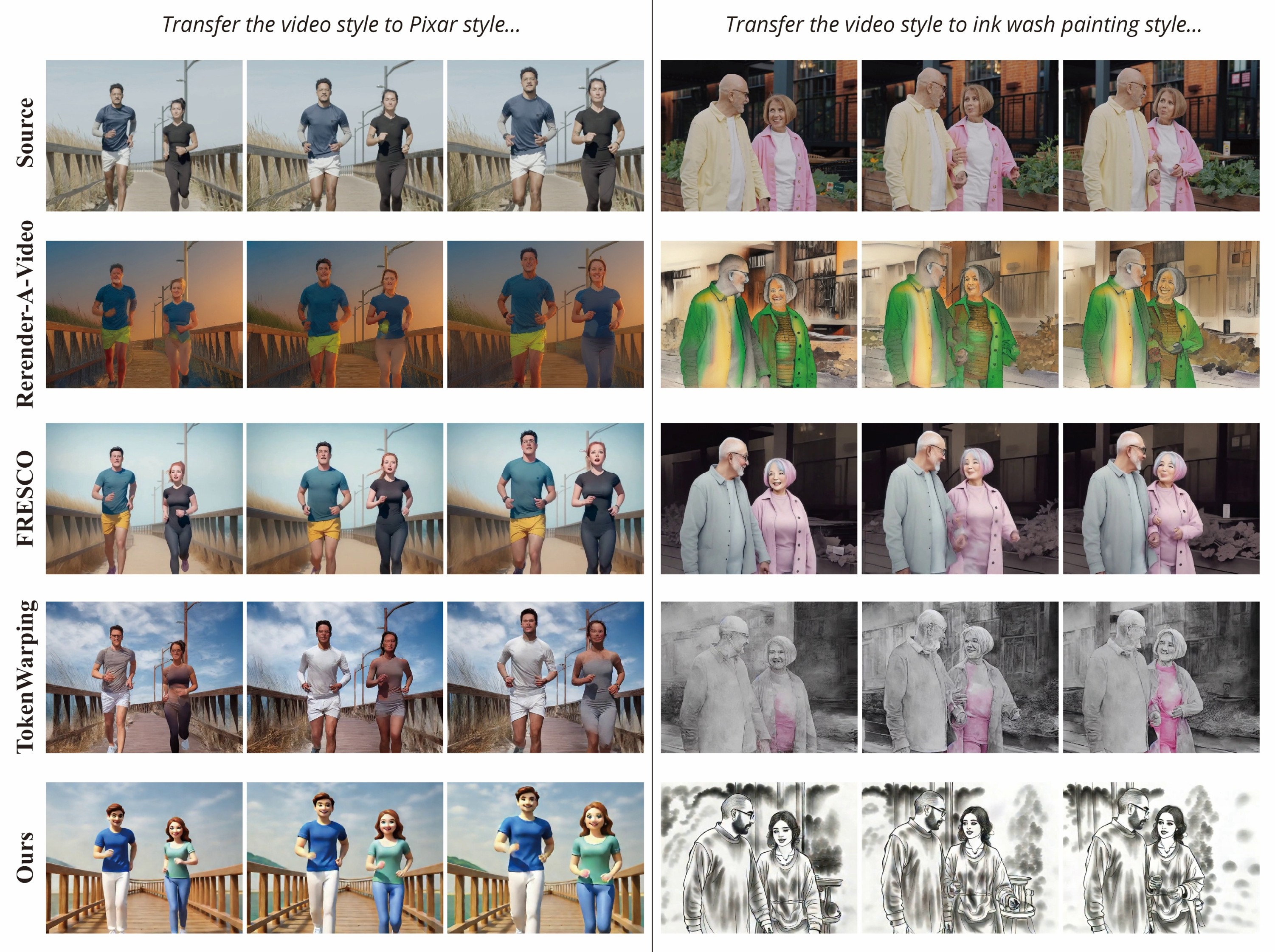}
  \caption{Visual comparison with TV2V stylization methods. Our method generates stylized videos that are more consistent with the text prompt and achieves superior visual quality compared to other methods.}
  \label{fig:tv2v}
\end{figure}

RTR-DiT is trained in three stages. First, we fine-tune a teacher model with bidirectional temporal modeling on our constructed dataset for 10,000 steps. Next, we initialize a few-step student model using a fixed timestep sequence $\{s_1,s_k,\cdots , s_K\}$ and train it for 10,000 steps to obtain a strong initialization. Finally, we apply post-training distillation to obtain the final model, trained for 5,000 steps. In our experiments, we set $K=2$, which provides the best trade-off between generation quality and inference speed. During training, we jointly train on a mixture of TV2V and RV2V samples.

\setlength{\tabcolsep}{5pt}
\begin{table}[b]
  \caption{Quantitative comparison results. Top: TV2V methods. Bottom: RV2V methods. Our method demonstrates strong performance across all metrics in both TV2V and RV2V settings.}
  \label{tab:qc}
  \centering
  \begin{tabular}{@{}lccccc@{}}
    \toprule
    Methods           & CLIP-T/CSD-Score          & CLIP-F          &   AQ            &   IQ            & Time(min)      \\
    \midrule
    Rerender-A-Video  &     0.2272                & 0.9883          & 0.5730          & 0.7012          & 4.5          \\
    FRESCO            &     0.1970                & 0.9907          & 0.6060          & 0.6834          & 7.3          \\
    TokenWarping      &     0.2122                & 0.9857          & 0.6093          & \textbf{0.7324} & 3.2          \\
    Ours              &     \textbf{0.2585}       &\textbf{0.9914}  & \textbf{0.6302} & 0.7279          & \textbf{0.12}  \\
    \hline
    VACE              &     0.4046                & 0.9905          & 0.6012          &          0.7165 & 27.6          \\
    StyleMaster       &     0.4275                & 0.9892          & 0.6536          & 0.7247          & 28.5          \\
    Gen-4 Aleph       &     \textbf{0.8312}       & \textbf{0.9925} & 0.6336          & \textbf{0.7378} & 3.5          \\
    Ours              &     0.7958                & 0.9909          & \textbf{0.6714} & 0.7321          & \textbf{0.12}  \\
  \bottomrule
  \end{tabular}
\end{table}

\subsubsection{Evaluation dataset.} 

We collect 50 videos from the Pexels website\cite{pexels} to build an evaluation benchmark. In TV2V mode, we prepare 10 diverse stylization prompts (e.g., ink wash painting and Minecraft styles) and evaluate the task using random text–video pairings from the benchmark. In the RV2V mode, to simplify the setup and enable standardized comparison with other reference-guided methods, we extract the first frame of each video and apply random style transformations using kling\cite{kling} as the reference. More RV2V stylization results can be found in the supplementary materials. All videos are resized to 5-second clips with a resolution of 832$\times$640 and a frame rate of 24 fps. Except for the commercial models, all tests are conducted on a single H20 GPU.

\subsection{Comparison with Text-Guided Video Stylization Methods}

We compare our approach with three diffusion-based stylization methods: Rerender-A-Video\cite{yang2023rerender-a-video}, FRESCO\cite{yang2024fresco}, and TokenWarping\cite{zhu2025tokenwarping}. Unlike our instructional prompts, these three methods use descriptive prompts. We use Qwen3-VL\cite{bai2025qwen3-vl} to generate descriptive prompts for the stylized scene content. For quantitative evaluation, we use the following metrics: CLIP-T\cite{radford2021clip} to measure the text-video similarity via CLIP, CLIP-F to evaluate temporal consistency through intra-frame similarity, and AestheticQuality(AQ) and ImagingQuality(IQ) from VBench\cite{huang2024vbench} to assess video quality. Additionally, we also measure the generation time for each video.

\cref{tab:qc} presents a comparison of our method with three other text-guided video stylization approaches. The results show that RTR-DiT achieves the best performance on CLIP-T, indicating that our method better follows the text guidance. It also performs the best on CLIP-F, demonstrating superior temporal consistency compared to the other methods. Additionally, our method achieves the highest AestheticQuality and second-best ImagingQuality, indicating that the generated video quality is notably superior. In terms of generation time, our method significantly outperforms the other three approaches, highlighting the efficiency advantage of autoregressive models in streaming stylization.

\cref{fig:tv2v} shows a comparison of the visual effects of the four methods. When converting to the ink wash painting style, the other methods incorrectly alter the video’s color tones while maintaining the scene's structure, whereas our method accurately transforms the video into the ink wash style. When converting to Pixar style, our method generates character designs that are closer to Pixar animation characters compared to the other methods. These results demonstrate the significant advantages of our method in text-guided video stylization.

\subsection{Comparison with Reference-Guided Video Stylization Methods}

\begin{figure}[t]
  \centering
  \includegraphics[width=0.8\textwidth]{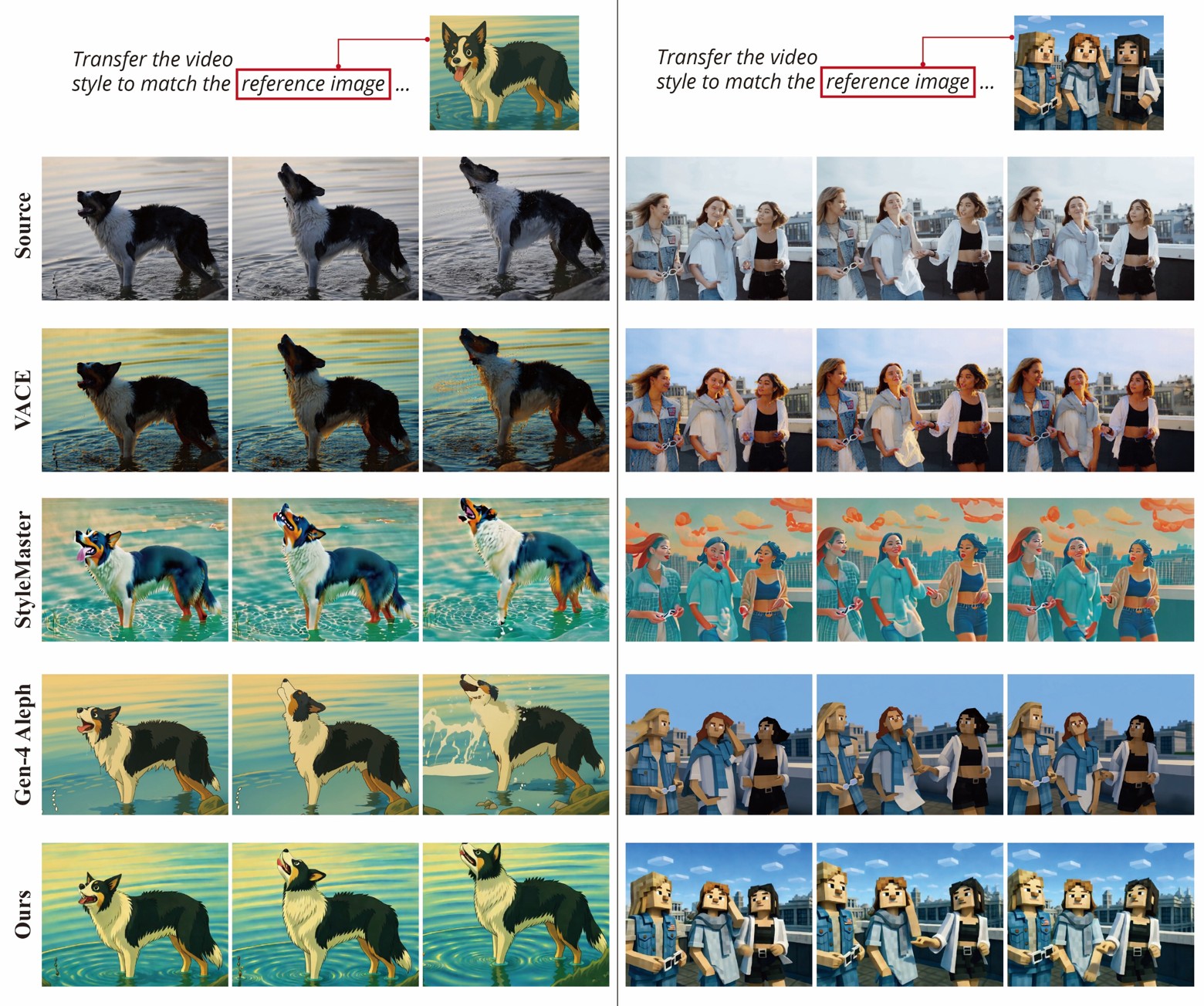}
  \caption{Visual comparison with RV2V stylization methods. Our method achieves performance comparable to commercial models under real-time inference conditions.}
  \label{fig:rv2v}
\end{figure}

We selected two DiT-based methods: VACE\cite{jiang2025vace} and StyleMaster\cite{ye2025stylemaster}, as well as the commercial model Gen-4 Aleph\cite{gen-4-aleph}, for comparison. In the evaluation metrics, we use CSD-Score\cite{somepalli2024csd-score} to replace CLIP-T, which measures the similarity between reference images and video frames.

\cref{tab:qc} presents a comparison of our method with three other reference-guided video stylization approaches. Our method achieves the best results in CLIP-F and AestheticQuality, and second-best in CSD-Score and ImagingQuality, just behind the commercial model Gen4-Aleph. This indicates that RTR-DiT outperforms existing open-source methods in reference-guided character stylization and is competitive with commercial models. 

\cref{fig:rv2v} presents a qualitative comparison of the four methods. VACE, when generating stylized videos guided by reference images, only adjusts the image brightness and fails to capture the style features effectively. The stylized videos generated by StyleMaster show significant visual style differences compared to the reference image. Both Runway and our method generate high-quality stylized results, but our results are closer to the reference image and more stable.

In addition, VACE and StyleMaster struggle with long inference times, which limits their applicability in real-time scenarios. Although the commercial model Gen-4 Aleph achieves shorter inference times than the former two methods, users still have to endure long queuing times in practice. In contrast, our method maintains fast generation while still producing high-quality results.

\begin{figure}[t]
\centering
\begin{subfigure}{0.49\linewidth}
    \centering
    \includegraphics[width=\linewidth]{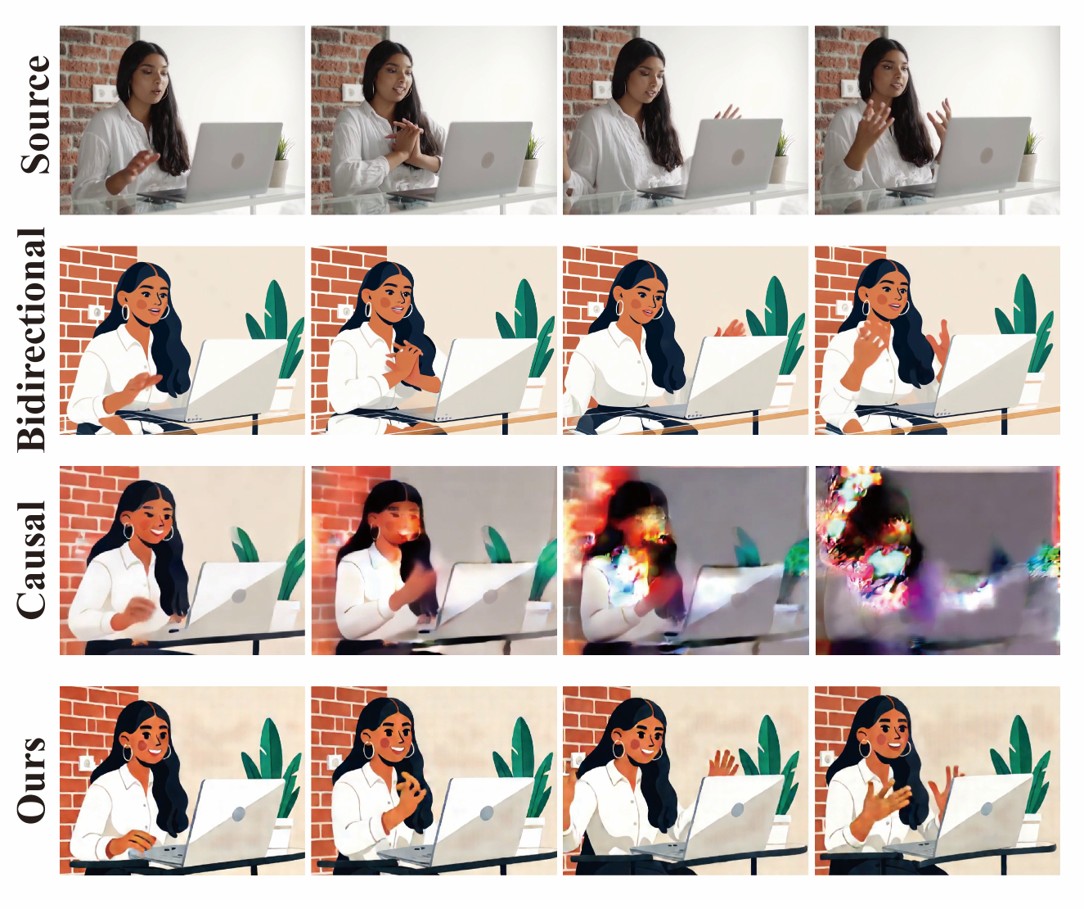}
    \caption{Visual comparison of results at different training stages. The model without post-training fails to generate stable long videos, whereas our model produces results comparable in quality to the bidirectional model.}
    \label{fig:train-stage}
\end{subfigure}
\hfill
\begin{subfigure}{0.49\linewidth}
    \centering
    \includegraphics[width=\linewidth]{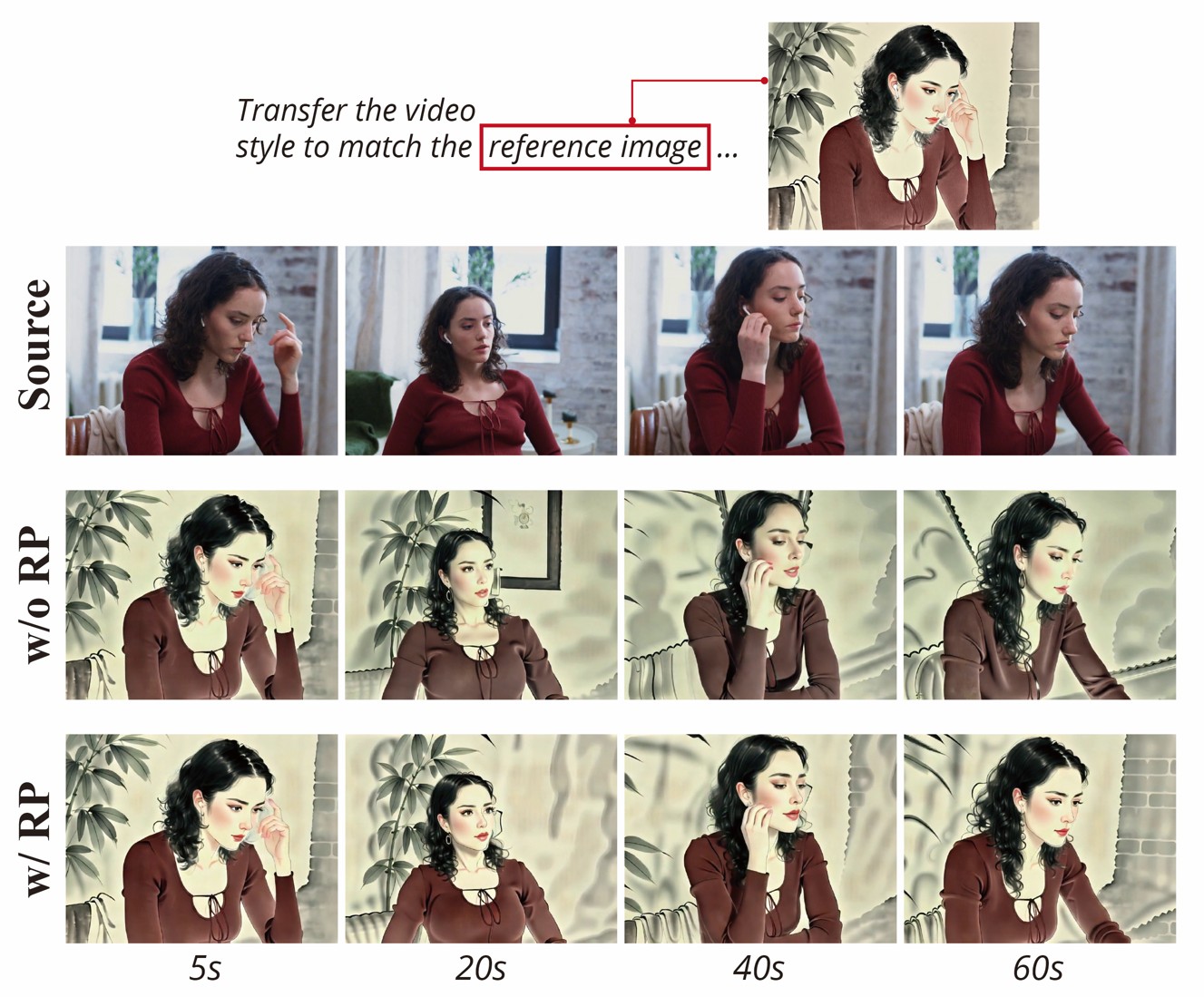}
    \caption{The reference-preserving (RP) update strategy effectively stabilizes long video stylization, ensuring that newly generated frames remain consistent with the reference image's style.}
    \label{fig:rp}
\end{subfigure}

\caption{Qualitative comparison results of ablation studies.}
\label{fig:ablation}
\end{figure}

We first evaluate the effectiveness of post-training by comparing the performance of models obtained at different training stages. The corresponding quantitative metrics and visual comparisons are shown in \cref{tab:ablation} and \cref{fig:ablation}. The bidirectional teacher model achieves strong performance across multiple metrics and produces high-quality visual results. In contrast, the causal model without post-training shows a noticeable drop in quantitative metrics, and the quality of generated frames gradually degrades as inference proceeds. After post-training, our model achieves results that are close to or even surpass the teacher model on several metrics, while producing visual results comparable to those of the teacher model.

We further validate the effectiveness of the proposed reference-preserving (RP) KV cache update strategy through qualitative evaluation. As shown in the figure, without fixing the reference tokens at the front of the cache, the generated frames gradually degrade during inference and drift away from the style of the reference image. In contrast, our method enables the model to generate videos with stable and consistent stylization.

\setlength{\tabcolsep}{5pt}
\begin{table}[t]
  \caption{Quantitative comparison across different training stages. Top: TV2V mode. Bottom: RV2V mode. After post-training, our method not only achieves fast inference but also shows significant improvements across all metrics. }
  \label{tab:ablation}
  \centering
  \begin{tabular}{@{}lccccc@{}}
    \toprule
    Methods           & CLIP-T/CSD-Score          & CLIP-F          &   AQ            &   IQ            & Time(min)      \\
    \midrule
    Bidirectional     &     0.2463                & 0.9931          & 0.6514          & 0.7406          & 1.43          \\
    Causal            &     0.2284                & 0.9948          & 0.5446          & 0.6380          & 0.12          \\
    Ours              &     0.2585                & 0.9914          & 0.6302          & 0.7279          & 0.12         \\
    \hline
    Bidirectional     &     0.8050                & 0.9930          & 0.6583          & 0.7144          & 1.43        \\
    Causal            &     0.6002                & 0.9956          & 0.5897          & 0.6021          & 0.12          \\
    Ours              &     0.7958                & 0.9909          & 0.6714          & 0.7321          & 0.12          \\
  \bottomrule
  \end{tabular}
\end{table}

\subsection{Long Video Stylization and Applications}

\begin{figure}[t]
  \centering
  \includegraphics[width=0.8\textwidth]{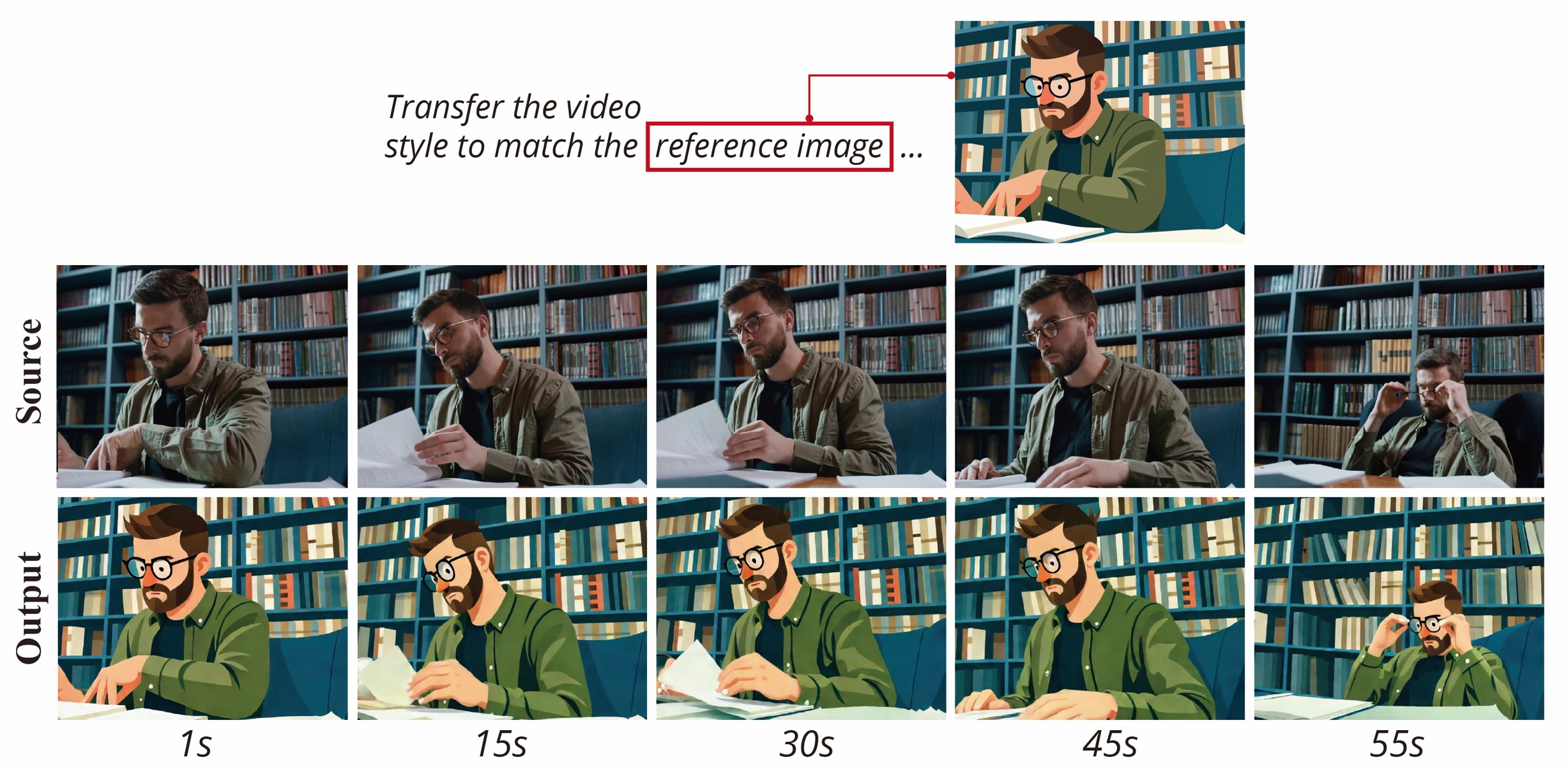}
  \caption{Long video stylization example. RTR-DiT maintains stable stylization over extended durations under streaming input.}
  \label{fig:long-video}
\end{figure}

\begin{figure}[t]
  \centering
  \includegraphics[width=0.8\textwidth]{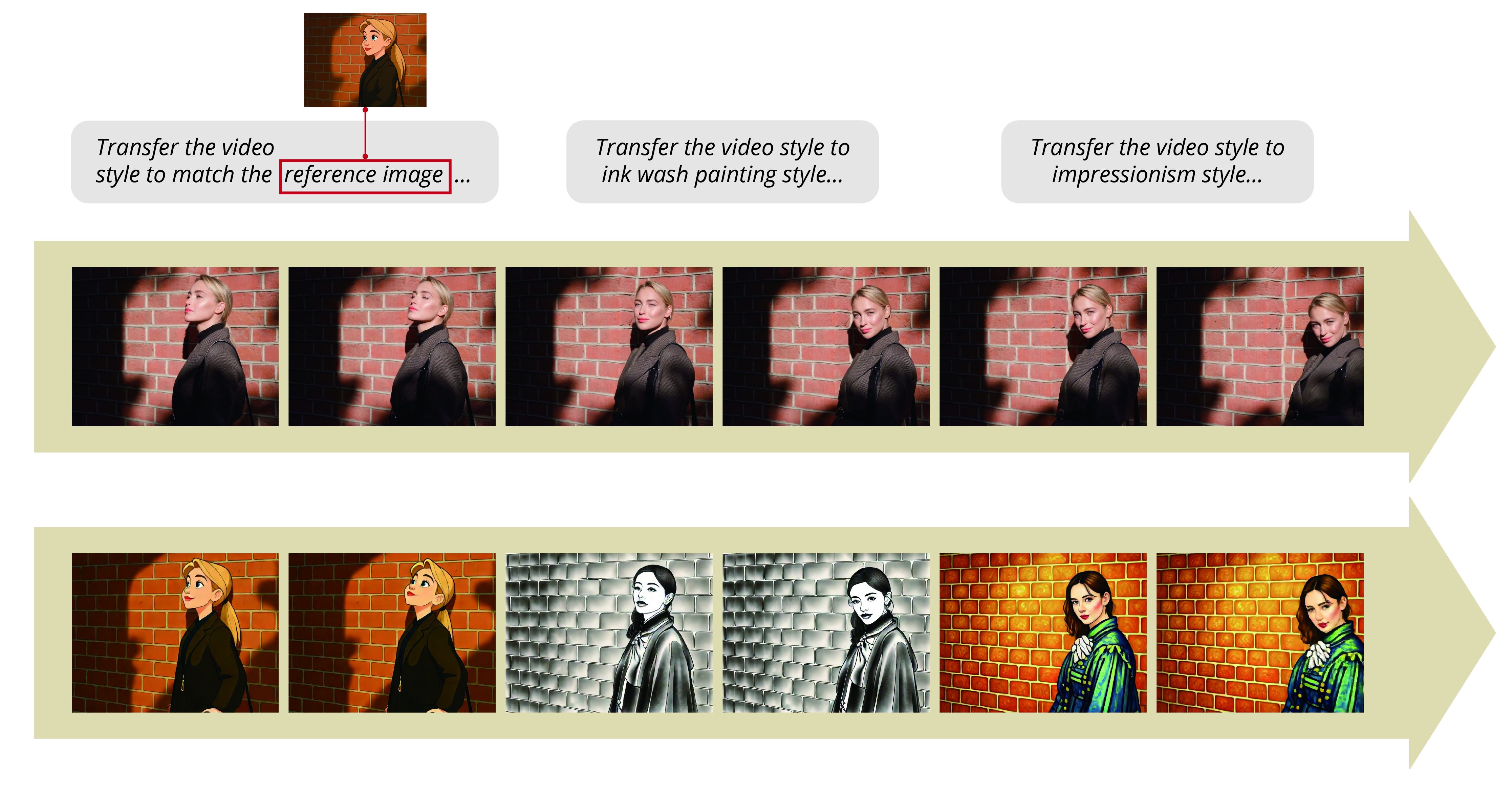}
  \caption{Text/reference-guided switching example. Our model can switch the guidance conditions in real-time.}
  \label{fig:switch}
\end{figure}

We collect several long videos from real-world scenes to demonstrate the ability of RTR-DiT in stable long-video stylization. As shown in \cref{fig:long-video}, our model is capable of maintaining stylization over long videos of approximately 1-minute duration, with the generated video consistently matching the style of the reference image. This is attributed to the enhancement of autoregressive DiT capabilities through post-training and the improvement in inference performance brought by the reference-preserving update strategy. More long-video stylization examples can be found in the supplementary materials.

We also demonstrate the ability to interactively switch the style of a video in real time by modifying the text or reference. As shown in Figure 8, the video is initially guided by an animated reference image, followed by a switch to a text-based prompt for an ink painting style, and finally, another text-based prompt is used to switch the style to Impressionism. RTR-DiT enables style changes under different guidance conditions, and the generated results remain faithful to the guidance conditions while maintaining high quality and consistency. This demonstrates the significant research potential of our method in real-time immersive applications such as VR/AR. Based on RTR-DiT, we have developed a real-time stylization application that uses a smartphone as the front end, streaming the captured video to a network deployed on a server and outputting the stylized results in real time. Further details and additional style-switching examples can be found in the supplementary materials.

%% file: 5_conclusion.tex
\section{Conclusion}

In this paper, we propose RTR-DiT, a real-time video stylization framework based on DiT that enables streaming, high-quality, and spatiotemporally consistent video stylization through both text and reference image guidance. To achieve this, we construct a high-quality long video stylization dataset containing text-video and reference-video pairs, and train an autoregressive generator through a staged approach. We first fine-tune a bidirectional DiT as the teacher model, then design a self-rollout causal DiT as the student model, and use the DMD algorithm for distillation training. Additionally, we introduce a reference-preserving KV cache update strategy to enable stable and consistent processing of long videos, while supporting real-time switching between prompts and reference images. Experimental results show that our method outperforms existing open-source approaches in both TV2V and RV2V modes in terms of quantitative metrics and visual quality, and is competitive with commercial models.